%% file: main.tex
\documentclass[10pt,twocolumn,letterpaper]{article}

\usepackage[pagenumbers]{cvpr} 
\usepackage[accsupp]{axessibility}  
\usepackage{booktabs}
\usepackage{multirow}
\input{preamble}

\definecolor{cvprblue}{rgb}{0.21,0.49,0.74}
\usepackage[pagebackref,breaklinks,colorlinks,allcolors=cvprblue]{hyperref}

\def\paperID{2138} 
\def\confName{CVPR}
\def\confYear{2025}

\title{HEIE: MLLM-Based Hierarchical Explainable AIGC Image \\ Implausibility Evaluator}

\author{
Fan Yang\textsuperscript{\rm 1,2,5}\footnotemark[1]
\quad
Ru Zhen\textsuperscript{\rm 3}\footnotemark[1]
\quad
Jianing Wang\textsuperscript{\rm 4}
\quad
Yanhao Zhang\textsuperscript{\rm 3}\footnotemark[2]
\quad
Haoxiang Chen\textsuperscript{\rm 3}
\\
Haonan Lu\textsuperscript{\rm 3}
\quad
Sicheng Zhao\textsuperscript{\rm 1,2} \footnotemark[3]
\quad
Guiguang Ding\textsuperscript{\rm 1,2} \footnotemark[3]
\\
\textsuperscript{\rm 1}Tsinghua University
\ 
\textsuperscript{\rm 2}BNRist
\ 
\textsuperscript{\rm 3}OPPO AI Center
\\ 
\textsuperscript{\rm 4}Peking University
\ 
\textsuperscript{\rm 5}Hangzhou Zhuoxi Institute of Brain and Intelligence
\\
{\tt\small
yfthu@outlook.com, 
\{zhenru1,zhangyanhao,luhaonan\}@oppo.com, 
}
\\
{\tt\small
schzhao@gmail.com, 
dinggg@tsinghua.edu.cn
}
}

\begin{document}
\maketitle

{
\renewcommand{\thefootnote}{\fnsymbol{footnote}}
\footnotetext[1]{Equal contribution. \quad$\dagger$ Project lead. \quad$\ddagger$ Corresponding authors.}
}

\def\abb{HEIE}
\def\abbdata{Expl-AIGI-Eval}

\begin{abstract}
AIGC images are prevalent across various fields, yet they frequently suffer from quality issues like artifacts and unnatural textures. Specialized models aim to predict defect region heatmaps but face two primary challenges: (1) lack of explainability, failing to provide reasons and analyses for subtle defects, and (2) inability to leverage common sense and logical reasoning, leading to poor generalization. Multimodal large language models (MLLMs) promise better comprehension and reasoning but face their own challenges: (1) difficulty in fine-grained defect localization due to the limitations in capturing tiny details, and (2) constraints in providing pixel-wise outputs necessary for precise heatmap generation.
To address these challenges, we propose HEIE: a novel MLLM-Based \textbf{H}ierarchical \textbf{E}xplainable Image \textbf{I}mplausibility \textbf{E}valuator. We introduce the CoT-Driven Explainable Trinity Evaluator, which integrates heatmaps, scores, and explanation outputs, using CoT to decompose complex tasks into subtasks of increasing difficulty and enhance interpretability. Our Adaptive Hierarchical Implausibility Mapper synergizes low-level image features with high-level mapper tokens from LLMs, enabling precise local-to-global hierarchical heatmap predictions through an uncertainty-based adaptive token approach.
Moreover, we propose a new dataset: \abbdata, designed to facilitate interpretable implausibility evaluation of AIGC images. Our method demonstrates state-of-the-art performance through extensive experiments. 
Our project is at \href{https://yfthu.github.io/HEIE/}{https://yfthu.github.io/HEIE/}.
\end{abstract}

\vspace{-15pt}

\input{introduction}

\input{relatedworks}

\input{method}

\input{experiment}

\section{Conclusion}
In summary, we introduce the CoT-Driven Explainable Trinity Evaluator, which guides LLM to generate the heatmap, score, and explanation for image implausibility. We propose an Adaptive Hierarchical Implausibility Mapper capable of enhancing the prediction of subtle implausibilities. Additionally, we present the \abbdata~dataset, which offers high-quality, explainable implausibility annotations. Our work facilitates users in understanding the image implausibility and improving the images. In the future, we aim to improve the reasoning efficiency and performance of the MLLM-based framework.

{
    \small
    \bibliographystyle{ieeenat_fullname}
    \bibliography{main}
}

\end{document}

%% file: preamble.tex
\usepackage{xcolor}


%% file: introduction.tex
\begin{figure}[ht]
\begin{center}
\includegraphics[width=1\linewidth]{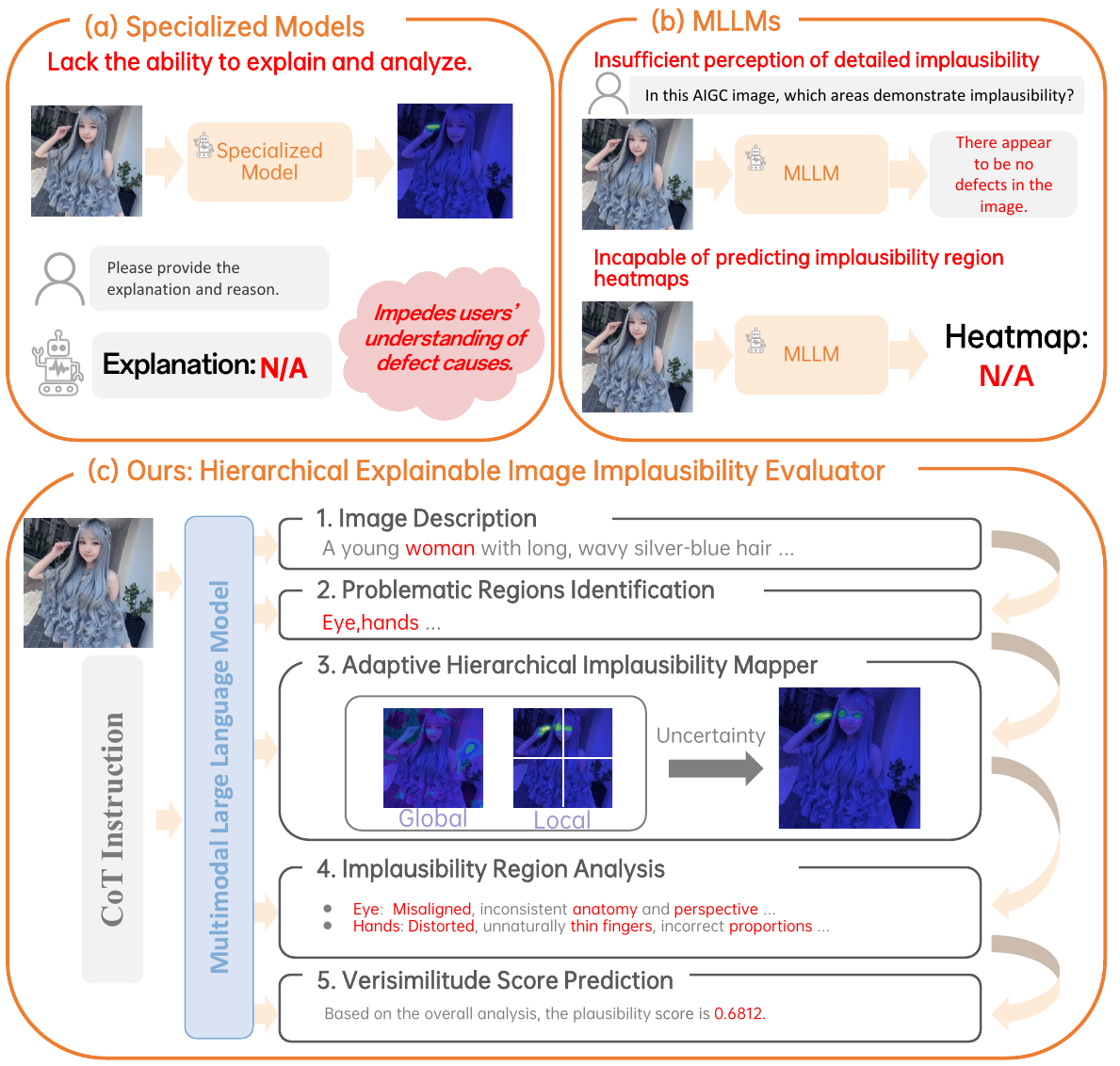}
\end{center}
\caption{(a) Specialized models lack the ability to explain and analyze subtle implausibility regions, hindering understanding for general users. (b) MLLMs struggle with precise localization of local defects and cannot directly output pixel-level implausibility areas. (c) Our CoT-Driven Explainable Trinity Evaluator can generate heatmaps, analyses, and scores. In our Adaptive Hierarchical Implausibility Mapper, local and global heatmaps are predicted separately, improving the localization of tiny implausibilities.}
\label{Fig:introduction}
\end{figure}
\section{Introduction}
Recently, Artificial Intelligence Generated Content (AIGC) has rapidly advanced, leading to the widespread use of AIGC images across various fields.
However, they often face quality issues such as artifacts, unnatural textures, and problematic regions. Accordingly, many studies focus on the evaluation of AIGC images \cite{imagereward,human_preference_score,pick-a-pic}. Current evaluations primarily yield scalar scores, simple 0-1 values, failing to identify specific defect regions, which hinders targeted improvements in generative models.

A few months ago, RichHF \cite{richHF} introduced the task of predicting defect region heatmaps and the RAHF model, which locates specific problematic areas. However, specialized models like RAHF have several limitations in the implausibility evaluation task:

\textbf{(1) Lack of explainability}. As shown in Fig. \ref{Fig:introduction} (a), implausibility regions in AIGC images can be complex, such as twisted fingers or overly large eye corners. Specialized small models fail to provide detailed explanations for these subtle defects, making it difficult for general users to understand and optimize the AIGC models.

\textbf{(2) Lack of general knowledge}. These small specialized models are trained on limited datasets and lack comprehensive world knowledge, common sense, and logical reasoning skills. As a result, they struggle to generalize to out-of-domain samples during practical usage.

 One approach to tackle the above issues is to use MLLMs~\cite{bai2023qwen,lu2024deepseek,glm2024chatglm}. MLLMs have rapidly advanced through large training data and excel in understanding, generating, and possessing extensive knowledge and common. However, directly applying vanilla MLLM to image implausibility evaluation remains challenging:

\textbf{(1) Fine-Grained Defect Localization}: While MLLMs excel in capturing global image structure, they struggle with pinpointing fine-grained defects. Tiny defect areas like the corners of eyes or eyebrows (Fig. \ref{Fig:introduction} (b)) are often neglected by strong MLLMs such as GPT-4o. 

\textbf{(2) Pixel-Wise Output Limitations}: General-purpose MLLMs handle various modalities (text, images, videos, etc.) but usually output text only, lacking the capability to produce pixel-level defect heatmaps (Fig. \ref{Fig:introduction} (b)).

To overcome the issues of specialized small models and MLLMs in image implausibility evaluation, we propose HEIE: an MLLM-based \textbf{H}ierarchical \textbf{E}xplainable image \textbf{I}mplausibility \textbf{E}valuator, which overcomes various problems as follows:

Firstly, to address the issues of insufficient interpretability, common sense, and logical reasoning in previous methods, we propose the CoT-Driven Explainable Trinity Evaluator.
By decomposing the complex task into a sequential task structure, progressing from simple to difficult, our Chain of Thought (CoT) framework harnesses the full potential of LLMs to conduct step-by-step reasoning for intricate implausibility assessments. Our trinity evaluator not only generates scores and heatmaps but also incorporates the analysis and explanation of the implausibility.
Crucially, through the integration of CoT~\cite{wei2022chain,kojima2022large,zhang2022automatic} and structural design, our heatmaps, scores, and explanations are synergistic and mutually reinforcing.

Secondly, to overcome limitations in precise implausibility localization and pixel-level heatmap generation, we introduce the Adaptive Hierarchical Implausibility Mapper. Our Implausibility Mapper, based on two-way cross-attention, combines low-level image features from ViT ~\cite{alexey2020image} with high-level special map tokens from LLMs to predict implausibility heatmaps. AIGC image defects manifest as either global anomalies, such as extra limbs on an animal, or local irregularities, like issues with individual fingers. So, we propose an uncertainty-driven adaptive algorithm that integrates both global and local tokens. Furthermore, the adaptive map token dynamically adjusts according to image resolution and aspect ratio, thereby ensuring both flexibility and optimal performance.

Lastly, current image implausibility datasets do not meet our explainable evaluation needs. We propose a new dataset: \abbdata\ (Explainable AIGC Image Implausibility Evaluation Dataset). Utilizing visual prompting, LLM free-form outputting, and in-context learning-based formatting, this dataset provides interpretable implausibility evaluations, including image descriptions, problematic region identifications, and issue analyses, aiding CoT-based reasoning for LLMs.

Our contributions are as follows:
\begin{enumerate}
    \item We are the first to achieve explainable image implausibility evaluation. Our CoT-Driven Explainable System guides the LLM to decompose complex problems into subproblems of increasing difficulty and mutually reinforcing, improving heatmaps, scores, and explanations. \\
    \item We are the first to use MLLM for image implausibility heatmap prediction, leveraging MLLM's commonsense, logical reasoning, and generalization capabilities.\\
    \item We propose the Adaptive Hierarchical Implausibility Mapper, employing dynamic special map tokens for uncertainty-based global-local heatmap predictions, enhancing the localization of subtle defects.
    \item We construct and publish a high-quality explainable AIGC image implausibility evaluation dataset. Our method achieves state-of-the-art performance on various datasets and tasks. Extensive comparisons and ablation studies validate the effectiveness of our approach.
\end{enumerate}

%% file: relatedworks.tex
\begin{figure*}[ht]
\begin{center}
\includegraphics[width=1\linewidth]{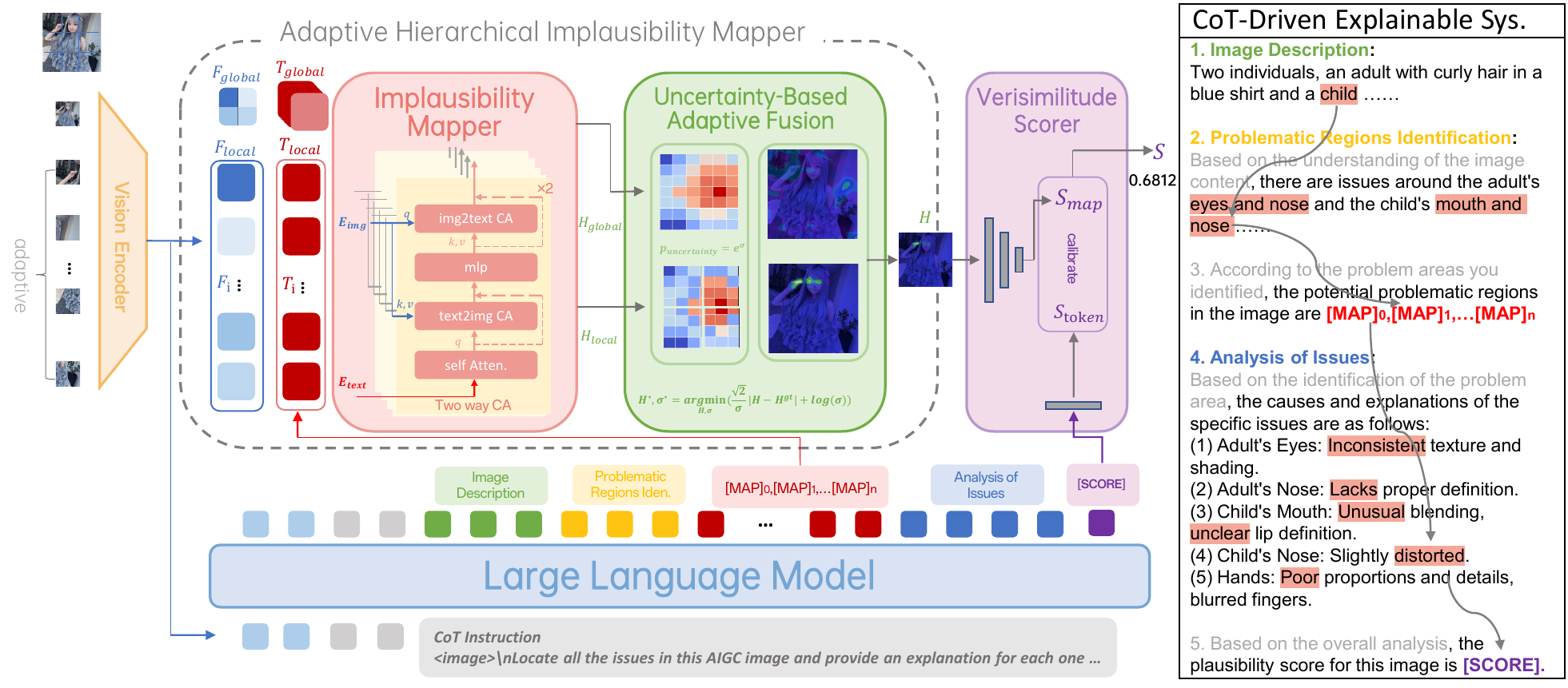}
\end{center}
\caption{The Implausibility Mapper processes a dynamic number of special \textbf{[MAP]} tokens and the image features to enhance detailed implausibility localization. We implement the Adaptive Hierarchical Implausibility Mapper through the local and global heatmaps and the uncertainty-based adaptive fusion. In the verisimilitude scorer, features from the heatmap and the special \textbf{[SCORE]} token are integrated for score prediction. Furthermore, our Cot-Driven Explainable System guides the LLM to decompose complex issues into progressive subproblems, facilitating the mutual enhancement of heatmap, analysis, and score prediction, thus improving explainability.}
\label{pipeline}
\end{figure*}

\section{Related Work}
\subsection{AI-Generated Images}
Advancements in text-to-image generation models have significantly impacted creative industries like art, design, and advertising. Stable Diffusion~\cite{rombach2022high}uses denoising processes to produce high-quality images. It is open-source, encouraging academic and industrial applications. DALL·E 3~\cite{dalle3} refines the generation of high-resolution, realistic images with detailed text integration.  MidJourney~\cite{midjourney}, by an independent lab, focuses on creating artistic and imaginative images, prioritizing aesthetics over realism. Recent works~\cite{li2024snapfusion} explore combining diffusion models with GANs, showing potential for enhancing image quality and diversity. However, AIGC images often contain many implausible elements that require careful evaluation.

\subsection{AIGC Image Evaluation}

Effective evaluation are required to optimize generative models. Many studies have released datasets to emulate human assessment~\cite{imagereward, human_preference_score, pick-a-pic,TIFA,AGIQA-3K}. Large-scale visual understanding models are now commonly used for reliable evaluation due to their strong semantic capabilities~\cite{peng2024dreambench++,wu2024comprehensive,you2023depicting}. Prior research has shown that MLLMs are proficient in both low-level visual evaluations~\cite{q-bench,q-instruct} and high-level semantic comprehension~\cite{A-Bench,yang2024llmi3d,shen2025llava,shen2025fastvid,shen2024tempme}. Some studies ~\cite{qalign, evalalign, VisualCritic,wu2025towards} activate the visual evaluation capabilities of multimodal MLLMs through data fine-tuning.

Existing models primarily aim to align their scores with human preferences~\cite{qalign,human_preference_score} or aesthetic evaluation ~\cite{aesbench,AesExpert}. 
These approaches fail to identify specific fine-grained problem areas within images, which hinders users and researchers from making targeted improvements.

Recently, HumanRefiner~\cite{fang2024humanrefiner} and RichHF ~\cite{richHF} have introduced tasks focused on detecting fine-grained problematic regions in images. HumanRefiner targets defects in AIGC-generated human images, while RichHF encompasses defects in various objects. However, these approaches rely on specialized small models to detect defect regions. These models are limited in common sense and knowledge, resulting in insufficient generalization capabilities in complex scenarios. Additionally, specialized small models lack interpretability and cannot provide detailed reasons or explanations ~\cite{angelov2021explainable,dovsilovic2018explainable} for the detected issues.

%% file: method.tex
\section{Method}
\subsection{Overview}
As shown in Fig. \ref{pipeline}, \abb~first employs an Adaptive Hierarchical Implausibility Mapper to generate an image implausibility region heatmap. Subsequently, a verisimilitude scorer is used to predict the image score. Furthermore, with the aid of a Co-driven Explainable System, we stimulated the potential of the LLM to achieve mutual enhancement and joint output of the heatmap, score, and explanation.

\newcommand{\Etext}{E_{\text{text}}}
\newcommand{\Eimg}{E_{\text{img}}}
\newcommand{\Himgtext}{C_{\text{text→img}}}
\newcommand{\Htextimg}{C_{\text{img→text}}}
\newcommand{\Etextpie}{E_{\text{text}^{'}}}
\newcommand{\Output}{O_{2way\_CA}}

\subsection{Adaptive Hierarchical Implausibility Mapper}
\label{Adaptive Hierarchical Implausibility Mapper}

\subsubsection{MLLM-Based Implausibility Mapper}
\label{MLLM-Based Implausibility Mapper}

To obtain fine-grained, precise image implausibility regions, pixel-level mask outputs are necessary. However, most MLLMs only provide text information and cannot produce heatmap predictions. To address this, we design the Image Implausibility Mapper module for heatmap prediction. We introduced a special map token \textbf{[MAP]} to extract implausibility region information from LLM. 
In LLM, the \textbf{[MAP]} token generates many feature layers, and its final linear layer feature is used to predict the next token for LLM.
We select this final LLM layer’s hidden state of the \textbf{[MAP]} token as the features $T$.
Then we use image features $F$ from the image encoder and \textbf{[MAP]} token features $T$ from the LLM for a two-way cross-attention calculation~\cite{sam,alexey2020image,vaswani2017attention}, as shown in Eq.~\ref{2waytrans}.

\begin{small}
\begin{equation}
\begin{aligned}
\label{2waytrans}
  &C_{\text{T→F}} = \text{CrossAtten}(\text{SelfAtten}(T), F, F) \\
  &T^{\prime} = \text{mlp}(\text{norm}(C_{\text{T→F}})) \\
  &C_{\text{F→T}} = \text{CrossAtten}(F, T^{\prime}, T^{\prime}) \\
  & \Output= \text{norm}(C_{\text{F→T}}) 
\end{aligned}
\end{equation}
\end{small}

To fuse the \textbf{[MAP]} token features $T$ with image features $F$, each input alternates as the query and value during cross-attention operations.
By stacking the above two-way cross-attention Eq. (\ref{2waytrans}) twice, we get the predicted heatmap $H$.

Data imbalance poses a challenge in heatmap prediction, with mean squared error (MSE) loss often leading to conservative and low-value predictions. And we simply adopt focal loss to overcome the data imbalance issue~\cite{focal}.

\subsubsection{Hierarchical Implausibility Mapper}

Implausibility detection is not the primary objective of mainstream general-purpose MLLMs. Defect localization often requires detailed information, including small defects like fingers and eyes, making it crucial to detect these nuances. We developed a method targeting both global and local defects by segmenting images into adaptive patches according to resolution ~\cite{chen2024far}. The image encoder processes the thumbnail and $N$ segmented patches, generating global features $F_g$ and local features $F_i$. The LLM then outputs $N$ local information-enhanced tokens, $\mathbf{[MAP]_{i}}$, for each patch. Tokens $T_i$ combine with local features $F_i$ and input into the Implausibility Mapper (see Sec. \ref{MLLM-Based Implausibility Mapper}), where cross-attention mechanisms generate local implausibility heatmaps $L_i$. These heatmaps are concatenated to form a complete heatmap:

\begin{equation}
H_{l}=F_{concat}(L_1, L_2, ..., L_N)
\end{equation}

To handle large implausibility regions and preserve semantics, we predict a global map token $T_g$. We utilize global features $F_g$ and the LLM token $T_g$ as inputs to the Implausibility Mapper to generate the heatmap $H_g$.

\subsubsection{Adaptive Implausibility Mapper}
AIGC images vary in resolution, aspect ratio, and content, necessitating adaptive strategies for handling implausibility.\\
\textbf{Adaptive special map token:} 
We propose an adaptive special map token to manage local information for AIGC images. Our hierarchical design adjusts the local token count based on image scale: more tokens for higher resolution, fewer for lower.
\\
\textbf{Adaptive Heatmap Fuser:}
Images can have both global and local defects. Global defects, like a person with three legs, require global features for implausibility prediction, while local errors, like incorrect hand or eye details, need local features.

We introduce an uncertainty-based local-global fusion strategy. We model Local Heatmap $H_l$ and Global Heatmap $H_g$ as Laplace distributions\footnote{The Laplace distribution is $f_X(x) = \frac{1}{2\lambda} \exp\left(-\frac{|x - \mu|}{\lambda}\right)$, where $\mu$ and $\lambda$ are parameters. The standard deviation $\sigma$ is $\sigma = \sqrt{2}\lambda$.} \cite{lu2021geometry}, with standard deviations $\sigma_l$ and $\sigma_g$.
During training, we minimize the terms $H$ and $\sigma$, where $H$ represents either $H_l$ or $H_g$, and $\sigma$ represents $\sigma_l$ or $\sigma_g$, respectively:
\begin{equation}
H^*, \sigma^* = \arg\min_{H,\sigma} \left( \frac{\sqrt{2}}{\sigma} |H - H^{gt}| + \log(\sigma) \right) 
\label{uncertatinty}
\end{equation}

During inference, we calculate confidence with Laplace-derived uncertainty and perform weighted summation using $p_{uncertainty} = e^{-\sigma}$ to obtain the final heatmap.

\begin{figure*}[htbp]
\begin{center}
\includegraphics[width=1\linewidth]
{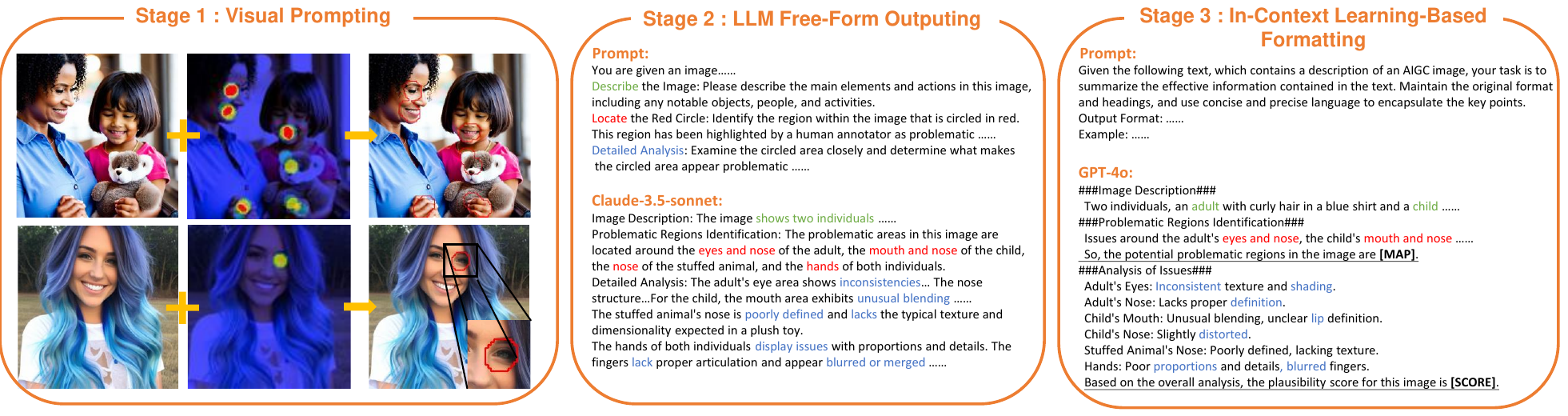}
\end{center}
\caption{\textbf{Three stages in \abbdata~dataset construction}: \textbf{Stage 1} - Visual Prompting: Defect regions are circled on images to aid Claude-3.5-sonnet in accurately locating problem areas. \textbf{Stage 2} - LLM Free-Form Output: Claude-3.5-sonnet generates free-form defect location and analysis. \textbf{Stage 3} - In-Context Learning-Based Formatting: GPT-4o is used for format standardization.}
\label{fig:dataset}
\end{figure*}

\subsection{CoT-Driven Explainable Trinity Evaluator}
\subsubsection{Verisimilitude Scorer}

Our model generates a heatmap of implausible regions and an overall verisimilitude score for comprehensive evaluation. LLMs struggle with numerical outputs, so we propose the Verisimilitude Scorer method to regress the score.

We define a special score token \textbf{[SCORE]} within the LLM, capturing the image's verisimilitude. The hidden state of the \textbf{[SCORE]} token is processed by a Feed-Forward Network (FFN) to regress an initial score, $S_{token}$.

Our experiments (Tab. \ref{ablheatmap2score}) show a strong correlation between the heatmap and verisimilitude score.
So we employ several convolution and FFN layers to get the heatmap score $S_{map}$ from the predicted heatmap.
The final score $S$ is derived by calibrating $S_{token}$ and $S_{map}$: $S = Calib (S_{token}, S_{map})$.
In our implementation, we explored various calibration functions, such as weighted summation and dynamic fusion. They exhibited similar performance.
\subsubsection{CoT-Driven Explainable System}
\label{CoT-Driven Explainable System}

In addition to generating the implausibility region heatmap and verisimilitude score, it is essential to analyze and explain these outputs for AIGC system improvement. We introduce the CoT-Driven Explainable Trinity Evaluator, depicted in Fig. \ref{pipeline}. The LLM is guided through a Chain of Thought (CoT) prompting process in five steps:

\textbf{(1) Image Description}: The LLM describes the image, capturing key elements in natural language.

\textbf{(2) Problematic Regions Identification}: Based on the description and the understanding of the image, the LLM identifies potential problematic areas for further analysis.

\textbf{(3) Special map token}: Based on the preceding Problematic Regions Identification, the LLM injects defect region information into the special map token for use by the implausibility mapper.

\textbf{(4) Analysis of Issues}: Based on the preceding implausibility localization, the LLM provides detailed text explanations for each region, including type, causes, and so on.

\textbf{(5) Special Score Token}:  Based on the comprehensive understanding, the LLM incorporates overall information into a special score token for use by our Verisimilitude Scorer.

This CoT-driven process breaks down the complex evaluation task into five progressively challenging and interrelated tasks, fully leveraging the LLM's potential.

\textbf{Trinity Implausibility Evaluator.}
Textual explanations, heatmaps, and scores synergize in image assessment. Texts describe image content, offering semantic context and guiding heatmap generation to highlight defect-prone areas. Heatmaps, using visual saliency, pinpoint critical regions, aiding scores in accurately quantifying defect levels. Conversely, scores refine heatmap focus, enhancing assessment precision. This interaction enables models to better understand and evaluate image issues.
\begin{figure*}[htbp]
\begin{center}
\includegraphics[width=1\linewidth]{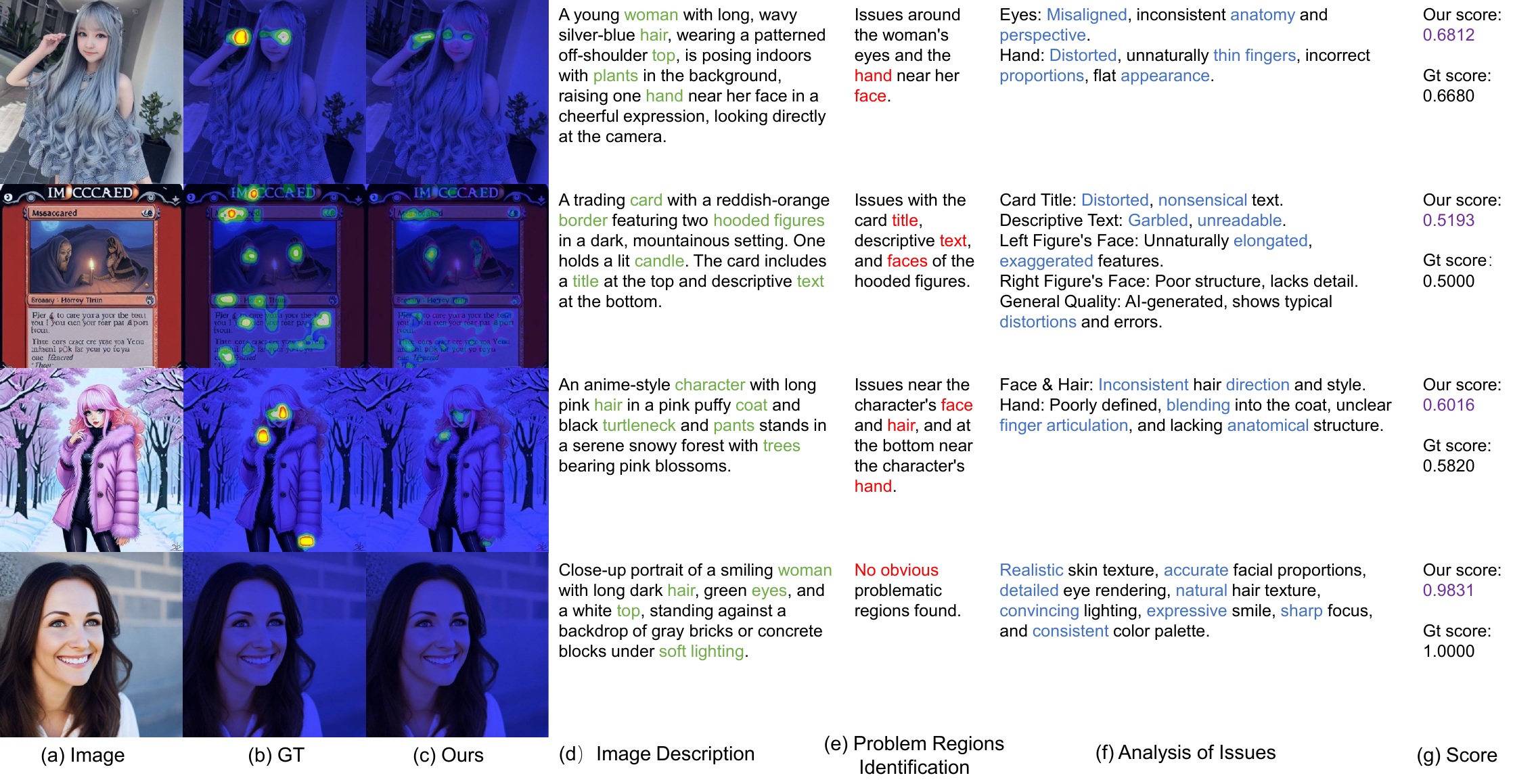}
\end{center}
\caption{\textbf{Model outputs of \abb.} \abb~not only predicts implausibility heatmaps but also provides image descriptions, problematic regions identification, analysis of issues, and score, achieving reliable and explainable implausibility evaluation. Note that for the last AIGC image, which has no evident defects, our model avoids false positives.}
\label{visable explain}
\end{figure*}

\begin{figure*}[ht]
\begin{center}
\includegraphics[width=1\linewidth]{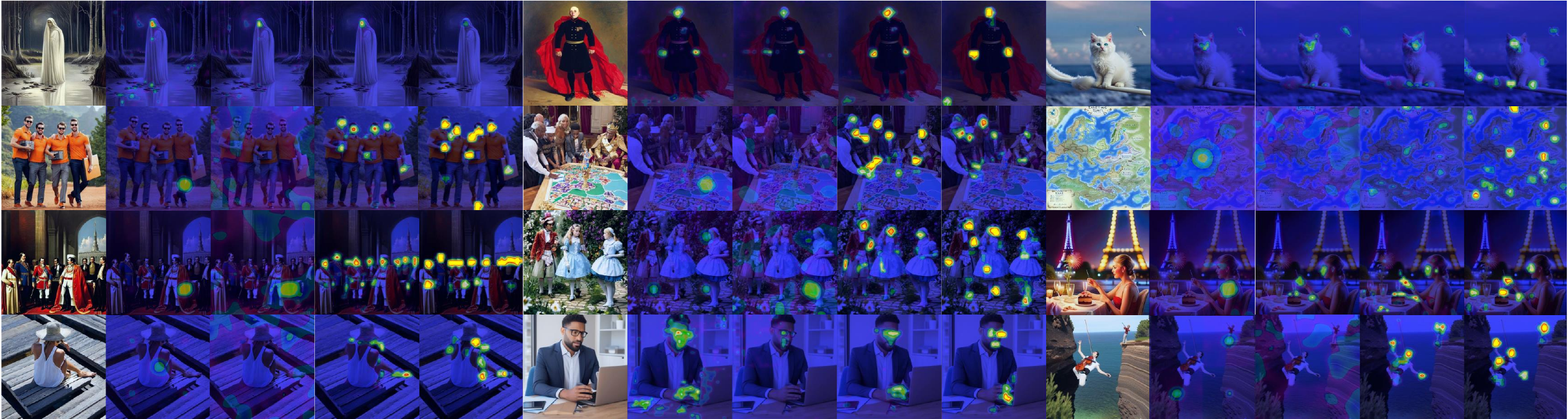}
\end{center}
\caption{\textbf{Comparison with Baselines.} Each set of images, from left to right, includes:  (a)  Input Image, (b) Output of InternViT-300M-448px, (c) Output of CLIP-ViT-Base-Patch16, (d) Output of ours \abb, (e) Ground Truth.}
\label{compare}
\end{figure*}

\subsection{\abbdata\ Dataset}

Existing AIGC image implausibility datasets, such as RichHF-18K~\cite{richHF} and AbHuman~\cite{fang2024humanrefiner}, include AIGC images and annotated defect regions but lack interpretability explanations. To address this, we introduce \abbdata: Explainable AIGC Image Implausibility Evaluation. We provide detailed analyses and explanations for each implausibility region. The annotation process is conducted in the following three stages, shown in Fig. \ref{fig:dataset}.

\textbf{(1) Visual Prompting}: We follow \cite{VisualGPT} to add visual prompts. Specifically, the original dataset provides masks of the defective regions for each image. We draw a red contour for each defective region, clearly highlighting these areas, as shown in Stage 1 in Fig. \ref{fig:dataset}. 

\textbf{(2) LLM Free-Form Outputting}: Using Claude-3.5-sonnet \cite{anthropic2024claude35sonnet}, we generate detailed analyses of the defect regions. Previous study \cite{speakfree} indicates that enforcing fixed-format outputs impairs the performance of LLM, so we allow the LLM to produce free-form outputs for analyzing defect causes. 

\textbf{(3) In-Context Learning-Based Formatting}: As shown in Stage-3 (Fig.\ref{fig:dataset}), we employ GPT-4o and in-context learning to format the defect analysis texts obtained in Stage 2. 

This hierarchical approach ensures comprehensive annotations. Each explanation includes image description, problematic regions identification, and issue analysis, supporting our CoT process in Sec. \ref{CoT-Driven Explainable System} and aiding CoT training.

Sensitive images related to pornography, privacy, animal protection, and political topics are filtered out to avoid ethical issues.

%% file: experiment.tex
\begin{figure}[htbp]
\begin{center}
\includegraphics[width=1\linewidth]{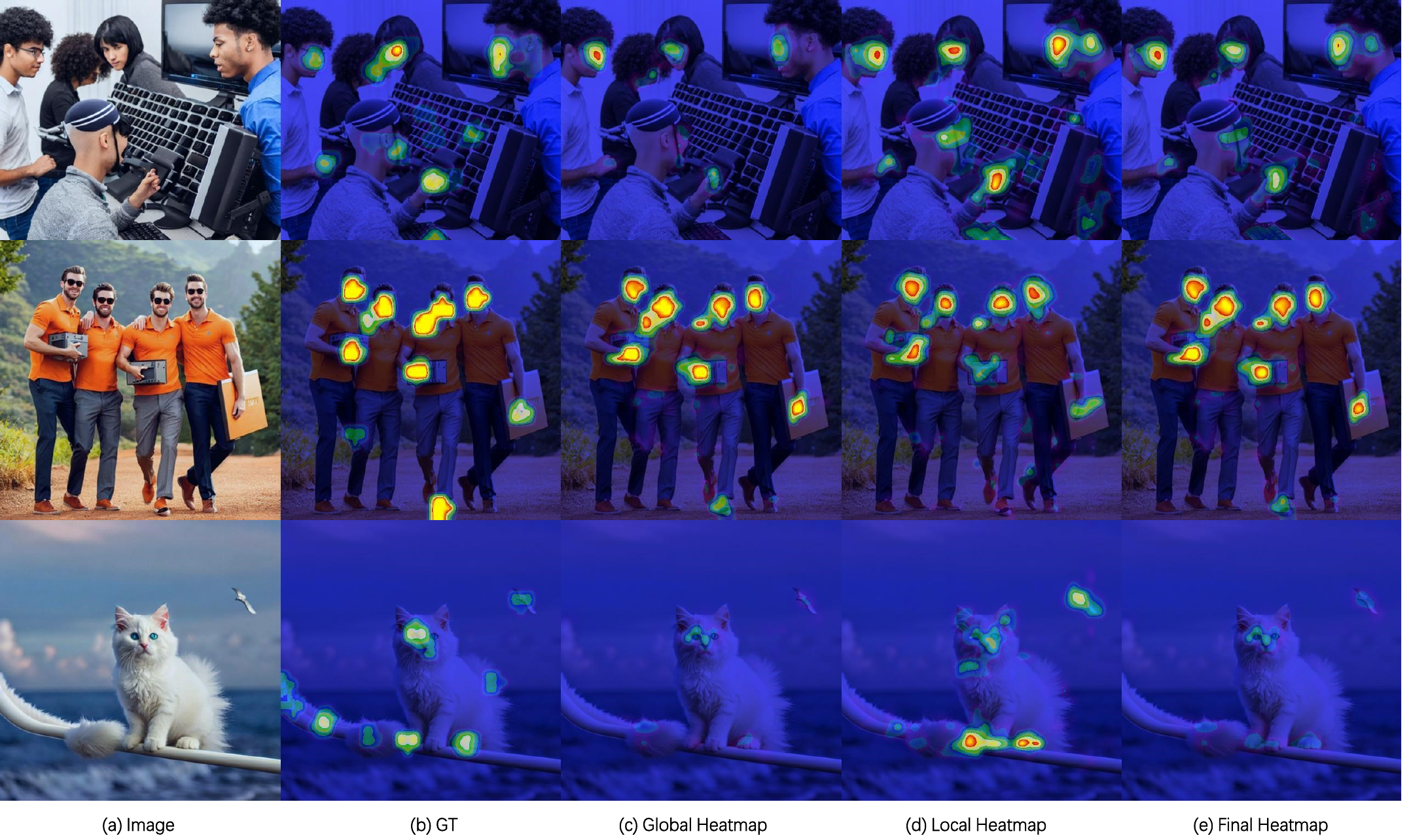}
\end{center}
\caption{\textbf{Results of Hierarchical Implausibility Mapper.} Columns (c) and (d) are the global and local heatmaps from our Hierarchical Implausibility Mapper. Column (e) illustrates the final heatmap after adaptively fusing the global and local heatmaps.}
\label{visable local}
\end{figure}

\begin{table*}[htbp]
\centering
\resizebox{\hsize}{!}{
\begin{tabular}{@{}c|cccccc|cc@{}}
\toprule
 & \multicolumn{6}{c|}{Heatmap Prediction} & \multicolumn{2}{c}{Plausibility Score} \\ \midrule
 & \multicolumn{1}{c|}{MSE (All Data)↓} & \multicolumn{1}{c|}{MSE (GT=0)↓} & KLD↓ & CC↑ & SIM↑ & AUC-Judd↑ & PLCC↑ & SRCC↑ \\
PickScore (off-the-shelf) & \multicolumn{1}{c|}{-} & \multicolumn{1}{c|}{-} & - & - & - & - & 0.010 & 0.028 \\
EVA-CLIP encoder (fine-tuned) & \multicolumn{1}{c|}{0.01614} & \multicolumn{1}{c|}{0.00512} & 2.835 & 0.350 & 0.082 & 0.549 & 0.157 & 0.143 \\
CLIP encoder (fine-tuned) & \multicolumn{1}{c|}{0.01437} & \multicolumn{1}{c|}{0.00425} & 2.462 & 0.251 & 0.122 & 0.747 & 0.390 & 0.378 \\
RAHF (multi-head) & \multicolumn{1}{c|}{0.01216} & \multicolumn{1}{c|}{0.00141} & 1.971 & 0.425 & 0.302 & 0.877 & 0.666 & 0.654 \\
RAHF (augmented prompt) & \multicolumn{1}{c|}{0.00920} & \multicolumn{1}{c|}{0.00095} & 1.652 & 0.556 & 0.409 & 0.913 & 0.693 & 0.681 \\
\abb~(ours) & \multicolumn{1}{c|}{\textbf{0.00825}} & \multicolumn{1}{c|}{\textbf{0.00014}} & \textbf{1.634} & \textbf{0.574} & \textbf{0.417} & \textbf{0.915} & \textbf{0.697} & \textbf{0.683} \\ \bottomrule
\end{tabular}
}
\caption{Comparison with State-of-the-Art on the RichHF-18K Dataset.}
\label{tab:richHF}
\end{table*}

\begin{table}[htbp]
\centering
\resizebox{\hsize}{!}{
\begin{tabular}{@{}l|lll@{}}
\toprule
                     & Perplexity ↓ & GPT-4o Eval ↑ & Human Eval ↑ \\ \midrule
Qwen2-VL-7B-Instruct~\cite{bai2023qwen} & 1.924209   & 1.910995  &  1.979058 \\    
DeepSeek-VL-7B-chat~\cite{lu2024deepseek}  & 1.794179   & 1.952880   &  1.883770 \\
InternVL2-8B~\cite{chen2024internvl}         & 1.456884   & 2.695288  &  2.603141  \\   
GLM-4V-9B~\cite{glm2024chatglm}            & 1.320043   & 2.486911  &  2.653403   \\
GPT-4o         &    \multicolumn{1}{c}{-}   & 3.828272  &  3.998953 \\
Claude-3.5-Sonnet  & \multicolumn{1}{c}{-}   & 3.938220  & 4.080628 \\
\abb~(ours)        &      \textbf{1.031390}      &    \textbf{4.582199}    &   \textbf{4.352880}  \\ \bottomrule
\end{tabular}
}
\caption{Performance of image implausibility explanations on our \abbdata~Dataset.}
\label{ppl}
\end{table}

\begin{table}[htbp]
\centering
\resizebox{\hsize}{!}{
\begin{tabular}{@{}c|c|c|cccc@{}}
\toprule
 & All Data & $GT=0$ & \multicolumn{4}{c}{$GT>0$} \\ \midrule
 & MSE ↓ & MSE ↓ & KLD↓ & CC↑ & SIM↑ & AUC-Judd↑ \\
InternViT\cite{chen2024internvl} & 0.07318 & 0.07248 & 3.515 & 0.019 & 0.091 & 0.524 \\
EVA-CLIP\cite{sun2024eva} & 0.00924 & 0.00207 & 3.226 & 0.582 & 0.095 & 0.607 \\
CLIP\cite{radford2021learning} & 0.00916 & 0.00920 & 1.953 & 0.244 & 0.154 & 0.636 \\
\abb~(ours) & \textbf{0.00510} & \textbf{0.00076} & \textbf{1.629} & \textbf{0.684} & \textbf{0.423} & \textbf{0.938} \\ \bottomrule
\end{tabular}
}
\caption{Results on AbHuman. All models are finetuned.}
\label{abhuman}
\end{table}

\begin{table*}[htbp]
\centering
\resizebox{\hsize}{!}{
\begin{tabular}{@{}c|cccc|c|cccc@{}}
\toprule
Train: RichHF-18K & All Data & $GT=0$ & \multicolumn{2}{c|}{$GT>0$} & Train: Abhuman & All Data & $GT=0$ & \multicolumn{2}{c}{$GT>0$} \\ \cmidrule(lr){2-5} \cmidrule(l){7-10} 
Test: AbHuman~\cite{fang2024humanrefiner} & MSE (All Data)↓ & MSE (GT=0)↓ & KLD↓ & CC↑ & Test: RichHF-18K & MSE (All Data)↓ & MSE (GT=0)↓ & KLD↓ & CC↑ \\ \midrule
InternViT-300MB~\cite{chen2024internvl} & 0.12610 & 0.12384 & 3.263 & -0.046 & InternViT-300MB & 0.25362 & 0.25170 & 10.547 & -0.017 \\
CLIP~\cite{radford2021learning} encoder & 0.01181 & 0.00492 & 2.730 & 0.212 & CLIP encoder & 0.01510 & 0.00384 & 3.448 & 0.015 \\
EVA-CLIP\cite{sun2024eva} & 0.01170 & 0.00562 & 3.225 & 0.482 & EVA-CLIP & 0.01432 & 0.00148 & 2.835 & 0.414 \\
\abb~(ours) & \textbf{0.00858} & \textbf{0.00298} & \textbf{1.796} & \textbf{0.506} & \abb~(ours) & \textbf{0.01091} & \textbf{0.00021} & \textbf{1.996} & \textbf{0.433} \\ \bottomrule
\end{tabular}
}
\caption{Performance of zero-shot domain generalization on the RichHF-18K and AbHuman dataset.}
\label{domain generalization}
\end{table*}

\section{Experiments}

\subsection{Experimental Setup}

\subsubsection{Implementation Details}
Our model is based on the InternVL-8B~\cite{chen2024internvl}. Our \abbdata\ dataset is constructed by expanding upon two existing datasets, RichHF-18K and AbHuman, by adding explanatory and analytical annotations. We follow the original train-validation split. We use Deepspeed to finetune the MLLM, with a learning rate of $3 \times 10^{-4}$, a warmup ratio of 0.03, and a batch size of 16. 

\subsubsection{Evaluation Metrics}
We follow the method in \cite{richHF} for evaluation. We report the MSE for all samples, including images with no problematic region. For images with implausibility regions, we provide evaluation metrics for the saliency heatmaps, such as KLD, AUC-Judd, SIM, and CC, as referenced in \cite{bylinskii2018different}. For the score prediction tasks, we report the Pearson linear correlation coefficient (PLCC) and Spearman rank correlation coefficient (SRCC) \cite{bylinskii2018different}. Additionally, since the RAHF model~\cite{richHF} is not open source, we provide its results reported in its paper.

\subsection{Comparison with State-of-the-art}

We evaluate our method on the RichHF-18K benchmark \cite{richHF}, as presented in Tab.~\ref{tab:richHF}. Conventional vision models typically lack physical common sense and struggle with high-level semantic understanding, complicating their ability to perceive and recognize content that goes against human intuition. In contrast, our method excels in accurately predicting heatmaps. Given the strong correlation between heatmap accuracy and image plausibility scores, our approach provides quantitative metrics that align with human evaluations. In comparison, our model achieved SOTA performance.

As shown in Tab. \ref{ppl}, we evaluate the performance of image implausibility explanations on our \abbdata~Dataset. We employ three evaluation methods to evaluate the models' text explanation capability:
(1) Compare the perplexity (ppl) of model outputs to ground truth; (2) Employ GPT-4o to compare the content similarity between the model outputs and the ground truth. 
(3) Conduct human evaluations of the model-generated explanations. 
Our carefully designed CoT approach significantly enhances our analysis and explanation capabilities for AIGC image implausibility. Our textual analysis outperforms both other open-source MLLMs and surpasses the general-purpose SOTA models like GPT-4o and Claude-3.5-sonnet. Across various textual explanation metrics, our model achieves SOTA performance.

Tab. \ref{abhuman} shows the results on the AbHuman dataset~\cite{fang2024humanrefiner}, which focuses more on the flaws in AI-generated images of humans. Baseline models underperform due to limited common sense and reasoning skills. In contrast, our model achieves SOTA results.

\subsection{Zero-shot Domain Generalization Results}

Tab. \ref{domain generalization} presents zero-shot cross-domain generalization experiments. We train on either the RichHF-18K or AbHuman datasets and perform zero-shot inference on the other. Cross-domain tasks \cite{yang2024geometry,he2022secret}, common in real-world applications \cite{wang2024yolov10,wang2024repvit,ding2021repvgg,ding2019acnet,yang2023gpro3d,yang2022ground}, test a model's fundamental world knowledge beyond domain-specific fitting \cite{hao2024quantized,wang2021lode}. As shown, smaller models suffer significant accuracy drops, losing defect assessment ability. In contrast, our model excels, leveraging MLLMs' strong common-sense knowledge and zero-shot capabilities.
\subsection{Ablation Study}

\begin{table}[htbp]
\centering
\resizebox{\hsize}{!}{
\begin{tabular}{@{}l|lllll|lll@{}}
\toprule
Exps & Global & Local & FW & LW & UncW & MSE↓ & KLD↓ & CC↑ \\ \midrule
(a) & \checkmark &  &  &  &  & 0.01071 & 1.950 & 0.502\\
(b) &  & \checkmark &  &  &  & 0.00980 & 1.921 & 0.504 \\
(c) & \checkmark & \checkmark & \checkmark &  &  & 0.00954 & 1.874 & 0.511 \\
(d) & \checkmark & \checkmark &  & \checkmark &  & 0.00873 & 1.680 & 0.557 \\
(e) & \checkmark & \checkmark &  &  & \checkmark & \textbf{0.00825} & \textbf{1.634} & \textbf{0.574} \\ \bottomrule
\end{tabular}
}
\caption{Ablation study for Adaptive Hierarchical Implausibility Mapper on RichHF-18K dataset.}
\label{ablahim}
\end{table}

\textbf{Ablation Study for Adaptive Hierarchical Implausibility Mapper.}
As shown in Tab. \ref{ablahim}, Exp. (a) uses only a special global map token and global image features for implausibility prediction, resulting in poor accuracy due to lost local details. Exp. (b) uses only a special local map token, preserving fine-grained image information and improving accuracy over Exp. (a). Exp. (c) combines global and local tokens with fixed weights, outperforming both singular approaches. Exp. (d) employs learnable weights, further improving results over Exp. (c). Exp. (e) uses an uncertainty-based adaptive strategy, dynamically weighting local and global heatmaps and achieving optimal performance.

\begin{table}[htbp]
\centering
\resizebox{\hsize}{!}{
\begin{tabular}{@{}c|ccc|cc@{}}
\toprule
Exps & Token & Heatmap & GT Heatmap & PLCC↑ & SRCC↑ \\ \midrule
(a) &  &  &  & 0.216 & 0.197 \\
(b) & \checkmark &  &  & 0.653 & 0.640 \\
(c) &  & \checkmark &  & 0.597 & 0.575 \\
(d) & \checkmark & \checkmark &  & 0.697 & 0.683 \\
(e) & \checkmark &  & \checkmark & \textcolor{gray}{0.743} & \textcolor{gray}{0.710} \\ \bottomrule
\end{tabular}
}
\caption{Ablation for Verisimilitude Scorer on RichHF-18K dataset.}
\label{ablheatmap2score}
\end{table}

\textbf{Ablation for Verisimilitude Scorer.}
Tab. \ref{ablheatmap2score} presents the results: Exp. (a) outputs scores in plain text via LLM, showing low accuracy due to LLM's numerical insensitivity. Exp. (b) uses the LLM's special score token \textbf{[SCORE]} for score prediction, while Exp. (c) uses only heatmap data. Both improve significantly over plain text. Exp. (d), our Verisimilitude Scorer, combines heatmap and \textbf{[SCORE]} token features with a calibration function for superior accuracy, validating its effectiveness. Exp. (e) further validates using ground truth heatmaps and \textbf{[SCORE]} token, achieving excellent score predictions, demonstrating the importance of accurate heatmaps.

\begin{table}[htbp]
\centering
\resizebox{\hsize}{!}{
\begin{tabular}{@{}c|ccc|cc@{}}
\toprule
 & \multicolumn{3}{c|}{Heatmap} & \multicolumn{2}{c}{Score} \\ \midrule
 & MSE↓ & KLD↓ & CC↑ & PLCC↑ & SRCC↑ \\
w/o CoT Text & 0.00913 & 1.689 & 0.553 & 0.669 & 0.658 \\
w/ CoT Text & 0.00825 & 1.634 & 0.574 & 0.697 & 0.683 \\
w/ GT CoT Text & \textcolor{gray}{0.00792} & \textcolor{gray}{1.601} & \textcolor{gray}{0.580} & \textcolor{gray}{0.701} & \textcolor{gray}{0.685} \\ \bottomrule
\end{tabular}
}
\caption{Ablation Study for CoT-Driven Explainable System.}
\label{ablcot}
\end{table}
\textbf{Ablation Study for CoT-Driven Explainable System.}
As shown in Tab. \ref{ablcot}, we evaluate our CoT Instruction. The first experiment without CoT Text (w/o CoT Text) reports results lacking our CoT Instruction, while the second (w/ CoT Text) shows outcomes with it. Comparing these, CoT Prompt significantly enhances heatmap and score predictions, demonstrating its effectiveness in guiding logical reasoning in LLMs. The third experiment (w/ GT CoT Text) uses ground truth CoT text labels, including image descriptions and problematic region identification, as instruction. These further boost prediction accuracy, affirming CoT's value in enhancing LLM reasoning.

\subsection{Visualization Analysis}
\quad\textbf{Evaluation Results of \abb:} As shown in Fig. \ref{visable explain}, by using CoT, \abb~implements comprehensive and explainable AIGC image implausibility evaluation.
Our model does not produce false positive predictions for defect-free images.
        
\textbf{Comparison with Baseline:} As Fig. \ref{compare} illustrates, predicting image implausibility heatmaps is challenging, and baselines perform poorly. Our model, however, shows excellent predictive performance. As well, we cannot obtain RAHF's prediction results due to the unavailability of RAHF's open model.

\textbf{Results of Hierarchical Implausibility Mapper:} As shown in Fig. \ref{visable local}, global heatmaps tend to predict obvious, coarse-grained regions, while local heatmaps focus on subtler, finer details. Their adaptive combination yields the final heatmap.

%% file: main.bbl
\begin{thebibliography}{58}
\providecommand{\natexlab}[1]{#1}
\providecommand{\url}[1]{\texttt{#1}}
\expandafter\ifx\csname urlstyle\endcsname\relax
  \providecommand{\doi}[1]{doi: #1}\else
  \providecommand{\doi}{doi: \begingroup \urlstyle{rm}\Url}\fi

\bibitem[Alexey(2020)]{alexey2020image}
Dosovitskiy Alexey.
\newblock An image is worth 16x16 words: Transformers for image recognition at scale.
\newblock \emph{arXiv preprint arXiv: 2010.11929}, 2020.

\bibitem[Angelov et~al.(2021)Angelov, Soares, Jiang, Arnold, and Atkinson]{angelov2021explainable}
Plamen~P Angelov, Eduardo~A Soares, Richard Jiang, Nicholas~I Arnold, and Peter~M Atkinson.
\newblock Explainable artificial intelligence: an analytical review.
\newblock \emph{Wiley Interdisciplinary Reviews: Data Mining and Knowledge Discovery}, 11\penalty0 (5):\penalty0 e1424, 2021.

\bibitem[Anthropic(2024)]{anthropic2024claude35sonnet}
Anthropic.
\newblock Claude 3.5 sonnet: A next-generation language model.
\newblock Online, 2024.
\newblock Version 3.5, Accessed: 2025-03-12.

\bibitem[Bai et~al.(2023)Bai, Bai, Yang, Wang, Tan, Wang, Lin, Zhou, and Zhou]{bai2023qwen}
Jinze Bai, Shuai Bai, Shusheng Yang, Shijie Wang, Sinan Tan, Peng Wang, Junyang Lin, Chang Zhou, and Jingren Zhou.
\newblock Qwen-vl: A frontier large vision-language model with versatile abilities.
\newblock \emph{arXiv preprint arXiv:2308.12966}, 2023.

\bibitem[Betker et~al.(2023)Betker, Goh, Jing, Brooks, Wang, Li, Ouyang, Zhuang, Lee, Guo, et~al.]{dalle3}
James Betker, Gabriel Goh, Li Jing, Tim Brooks, Jianfeng Wang, Linjie Li, Long Ouyang, Juntang Zhuang, Joyce Lee, Yufei Guo, et~al.
\newblock Improving image generation with better captions.
\newblock \emph{Computer Science. https://cdn. openai. com/papers/dall-e-3. pdf}, 2\penalty0 (3):\penalty0 8, 2023.

\bibitem[Bylinskii et~al.(2018)Bylinskii, Judd, Oliva, Torralba, and Durand]{bylinskii2018different}
Zoya Bylinskii, Tilke Judd, Aude Oliva, Antonio Torralba, and Fr{\'e}do Durand.
\newblock What do different evaluation metrics tell us about saliency models?
\newblock \emph{IEEE Transactions on Pattern Analysis and Machine Intelligence}, 41\penalty0 (3):\penalty0 740--757, 2018.

\bibitem[Chen et~al.(2022)Chen, Guo, Yi, Li, and Elhoseiny]{VisualGPT}
Jun Chen, Han Guo, Kai Yi, Boyang Li, and Mohamed Elhoseiny.
\newblock Visualgpt: Data-efficient adaptation of pretrained language models for image captioning.
\newblock In \emph{Proceedings of the IEEE/CVF Conference on Computer Vision and Pattern Recognition}, pages 18030--18040, 2022.

\bibitem[Chen et~al.(2024{\natexlab{a}})Chen, Wang, Tian, Ye, Gao, Cui, Tong, Hu, Luo, Ma, et~al.]{chen2024far}
Zhe Chen, Weiyun Wang, Hao Tian, Shenglong Ye, Zhangwei Gao, Erfei Cui, Wenwen Tong, Kongzhi Hu, Jiapeng Luo, Zheng Ma, et~al.
\newblock How far are we to gpt-4v? closing the gap to commercial multimodal models with open-source suites.
\newblock \emph{arXiv preprint arXiv:2404.16821}, 2024{\natexlab{a}}.

\bibitem[Chen et~al.(2024{\natexlab{b}})Chen, Wu, Wang, Su, Chen, Xing, Zhong, Zhang, Zhu, Lu, et~al.]{chen2024internvl}
Zhe Chen, Jiannan Wu, Wenhai Wang, Weijie Su, Guo Chen, Sen Xing, Muyan Zhong, Qinglong Zhang, Xizhou Zhu, Lewei Lu, et~al.
\newblock Internvl: Scaling up vision foundation models and aligning for generic visual-linguistic tasks.
\newblock In \emph{Proceedings of the IEEE/CVF Conference on Computer Vision and Pattern Recognition}, pages 24185--24198, 2024{\natexlab{b}}.

\bibitem[Ding et~al.(2019)Ding, Guo, Ding, and Han]{ding2019acnet}
Xiaohan Ding, Yuchen Guo, Guiguang Ding, and Jungong Han.
\newblock Acnet: Strengthening the kernel skeletons for powerful cnn via asymmetric convolution blocks.
\newblock In \emph{Proceedings of the IEEE/CVF international conference on computer vision}, pages 1911--1920, 2019.

\bibitem[Ding et~al.(2021)Ding, Zhang, Ma, Han, Ding, and Sun]{ding2021repvgg}
Xiaohan Ding, Xiangyu Zhang, Ningning Ma, Jungong Han, Guiguang Ding, and Jian Sun.
\newblock Repvgg: Making vgg-style convnets great again.
\newblock In \emph{Proceedings of the IEEE/CVF conference on computer vision and pattern recognition}, pages 13733--13742, 2021.

\bibitem[Do{\v{s}}ilovi{\'c} et~al.(2018)Do{\v{s}}ilovi{\'c}, Br{\v{c}}i{\'c}, and Hlupi{\'c}]{dovsilovic2018explainable}
Filip~Karlo Do{\v{s}}ilovi{\'c}, Mario Br{\v{c}}i{\'c}, and Nikica Hlupi{\'c}.
\newblock Explainable artificial intelligence: A survey.
\newblock In \emph{2018 41st International Convention on Information and Communication Technology, Electronics and Microelectronics (MIPRO)}, pages 0210--0215. IEEE, 2018.

\bibitem[Fang et~al.(2024)Fang, Yan, Guo, Han, Jiang, Xu, Liao, and Liang]{fang2024humanrefiner}
Guian Fang, Wenbiao Yan, Yuanfan Guo, Jianhua Han, Zutao Jiang, Hang Xu, Shengcai Liao, and Xiaodan Liang.
\newblock Humanrefiner: Benchmarking abnormal human generation and refining with coarse-to-fine pose-reversible guidance.
\newblock \emph{arXiv preprint arXiv:2407.06937}, 2024.

\bibitem[GLM et~al.(2024)GLM, Zeng, Xu, Wang, Zhang, Yin, Zhang, Rojas, Feng, Zhao, et~al.]{glm2024chatglm}
Team GLM, Aohan Zeng, Bin Xu, Bowen Wang, Chenhui Zhang, Da Yin, Dan Zhang, Diego Rojas, Guanyu Feng, Hanlin Zhao, et~al.
\newblock Chatglm: A family of large language models from glm-130b to glm-4 all tools.
\newblock \emph{arXiv preprint arXiv:2406.12793}, 2024.

\bibitem[Hao et~al.(2024)Hao, Ding, Feng, Yang, Chen, and Ding]{hao2024quantized}
Tianxiang Hao, Xiaohan Ding, Juexiao Feng, Yuhong Yang, Hui Chen, and Guiguang Ding.
\newblock Quantized prompt for efficient generalization of vision-language models.
\newblock \emph{arXiv preprint arXiv:2407.10704}, 2024.

\bibitem[He et~al.(2022)He, Shen, Guo, Ding, and Guo]{he2022secret}
Tao He, Leqi Shen, Yuchen Guo, Guiguang Ding, and Zhenhua Guo.
\newblock Secret: Self-consistent pseudo label refinement for unsupervised domain adaptive person re-identification.
\newblock In \emph{Proceedings of the AAAI conference on artificial intelligence}, pages 879--887, 2022.

\bibitem[Hu et~al.(2023)Hu, Liu, Kasai, Wang, Ostendorf, Krishna, and Smith]{TIFA}
Yushi Hu, Benlin Liu, Jungo Kasai, Yizhong Wang, Mari Ostendorf, Ranjay Krishna, and Noah~A Smith.
\newblock Tifa: Accurate and interpretable text-to-image faithfulness evaluation with question answering.
\newblock In \emph{Proceedings of the IEEE/CVF International Conference on Computer Vision}, pages 20406--20417, 2023.

\bibitem[Huang et~al.(2024{\natexlab{a}})Huang, Sheng, Yang, Yuan, Duan, Chen, Li, Lin, and Shi]{AesExpert}
Yipo Huang, Xiangfei Sheng, Zhichao Yang, Quan Yuan, Zhichao Duan, Pengfei Chen, Leida Li, Weisi Lin, and Guangming Shi.
\newblock Aesexpert: Towards multi-modality foundation model for image aesthetics perception.
\newblock In \emph{Proceedings of the 32nd ACM International Conference on Multimedia}, pages 5911--5920, 2024{\natexlab{a}}.

\bibitem[Huang et~al.(2024{\natexlab{b}})Huang, Yuan, Sheng, Yang, Wu, Chen, Yang, Li, and Lin]{aesbench}
Yipo Huang, Quan Yuan, Xiangfei Sheng, Zhichao Yang, Haoning Wu, Pengfei Chen, Yuzhe Yang, Leida Li, and Weisi Lin.
\newblock Aesbench: An expert benchmark for multimodal large language models on image aesthetics perception.
\newblock \emph{arXiv preprint arXiv:2401.08276}, 2024{\natexlab{b}}.

\bibitem[Huang et~al.(2024{\natexlab{c}})Huang, Zhang, Lu, Zha, Chen, and Guo]{VisualCritic}
Zhipeng Huang, Zhizheng Zhang, Yiting Lu, Zheng-Jun Zha, Zhibo Chen, and Baining Guo.
\newblock Visualcritic: Making lmms perceive visual quality like humans.
\newblock \emph{arXiv preprint arXiv:2403.12806}, 2024{\natexlab{c}}.

\bibitem[Kirillov et~al.(2023)Kirillov, Mintun, Ravi, Mao, Rolland, Gustafson, Xiao, Whitehead, Berg, Lo, et~al.]{sam}
Alexander Kirillov, Eric Mintun, Nikhila Ravi, Hanzi Mao, Chloe Rolland, Laura Gustafson, Tete Xiao, Spencer Whitehead, Alexander~C Berg, Wan-Yen Lo, et~al.
\newblock Segment anything.
\newblock In \emph{Proceedings of the IEEE/CVF International Conference on Computer Vision}, pages 4015--4026, 2023.

\bibitem[Kirstain et~al.(2023)Kirstain, Polyak, Singer, Matiana, Penna, and Levy]{pick-a-pic}
Yuval Kirstain, Adam Polyak, Uriel Singer, Shahbuland Matiana, Joe Penna, and Omer Levy.
\newblock Pick-a-pic: An open dataset of user preferences for text-to-image generation.
\newblock \emph{Advances in Neural Information Processing Systems}, 36:\penalty0 36652--36663, 2023.

\bibitem[Kojima et~al.(2022)Kojima, Gu, Reid, Matsuo, and Iwasawa]{kojima2022large}
Takeshi Kojima, Shixiang~Shane Gu, Machel Reid, Yutaka Matsuo, and Yusuke Iwasawa.
\newblock Large language models are zero-shot reasoners.
\newblock \emph{Advances in Neural Information Processing Systems}, 35:\penalty0 22199--22213, 2022.

\bibitem[Li et~al.(2023)Li, Zhang, Wu, Sun, Min, Liu, Zhai, and Lin]{AGIQA-3K}
Chunyi Li, Zicheng Zhang, Haoning Wu, Wei Sun, Xiongkuo Min, Xiaohong Liu, Guangtao Zhai, and Weisi Lin.
\newblock Agiqa-3k: An open database for ai-generated image quality assessment.
\newblock \emph{IEEE Transactions on Circuits and Systems for Video Technology}, 2023.

\bibitem[Li et~al.(2024)Li, Wang, Jin, Hu, Chemerys, Fu, Wang, Tulyakov, and Ren]{li2024snapfusion}
Yanyu Li, Huan Wang, Qing Jin, Ju Hu, Pavlo Chemerys, Yun Fu, Yanzhi Wang, Sergey Tulyakov, and Jian Ren.
\newblock Snapfusion: Text-to-image diffusion model on mobile devices within two seconds.
\newblock \emph{Advances in Neural Information Processing Systems}, 36, 2024.

\bibitem[Liang et~al.(2024)Liang, He, Li, Li, Klimovskiy, Carolan, Sun, Pont-Tuset, Young, Yang, et~al.]{richHF}
Youwei Liang, Junfeng He, Gang Li, Peizhao Li, Arseniy Klimovskiy, Nicholas Carolan, Jiao Sun, Jordi Pont-Tuset, Sarah Young, Feng Yang, et~al.
\newblock Rich human feedback for text-to-image generation.
\newblock In \emph{Proceedings of the IEEE/CVF Conference on Computer Vision and Pattern Recognition}, pages 19401--19411, 2024.

\bibitem[Lu et~al.(2024)Lu, Liu, Zhang, Wang, Dong, Liu, Sun, Ren, Li, Yang, et~al.]{lu2024deepseek}
Haoyu Lu, Wen Liu, Bo Zhang, Bingxuan Wang, Kai Dong, Bo Liu, Jingxiang Sun, Tongzheng Ren, Zhuoshu Li, Hao Yang, et~al.
\newblock Deepseek-vl: towards real-world vision-language understanding.
\newblock \emph{arXiv preprint arXiv:2403.05525}, 2024.

\bibitem[Lu et~al.(2021)Lu, Ma, Yang, Zhang, Liu, Chu, Yan, and Ouyang]{lu2021geometry}
Yan Lu, Xinzhu Ma, Lei Yang, Tianzhu Zhang, Yating Liu, Qi Chu, Junjie Yan, and Wanli Ouyang.
\newblock Geometry uncertainty projection network for monocular 3d object detection.
\newblock In \emph{Proceedings of the IEEE/CVF International Conference on Computer Vision}, pages 3111--3121, 2021.

\bibitem[{MidJourney}(2023)]{midjourney}
{MidJourney}.
\newblock Midjourney: Ai-based image synthesis tool.
\newblock \url{https://www.midjourney.com}, 2023.
\newblock Accessed: 2023-10-01.

\bibitem[Peng et~al.(2024)Peng, Cui, Tang, Qi, Dong, Bai, Han, Ge, Zhang, and Xia]{peng2024dreambench++}
Yuang Peng, Yuxin Cui, Haomiao Tang, Zekun Qi, Runpei Dong, Jing Bai, Chunrui Han, Zheng Ge, Xiangyu Zhang, and Shu-Tao Xia.
\newblock Dreambench++: A human-aligned benchmark for personalized image generation.
\newblock \emph{arXiv preprint arXiv:2406.16855}, 2024.

\bibitem[Radford et~al.(2021)Radford, Kim, Hallacy, Ramesh, Goh, Agarwal, Sastry, Askell, Mishkin, Clark, et~al.]{radford2021learning}
Alec Radford, Jong~Wook Kim, Chris Hallacy, Aditya Ramesh, Gabriel Goh, Sandhini Agarwal, Girish Sastry, Amanda Askell, Pamela Mishkin, Jack Clark, et~al.
\newblock Learning transferable visual models from natural language supervision.
\newblock In \emph{International Conference on Machine Learning}, pages 8748--8763. PMLR, 2021.

\bibitem[Rombach et~al.(2022)Rombach, Blattmann, Lorenz, Esser, and Ommer]{rombach2022high}
Robin Rombach, Andreas Blattmann, Dominik Lorenz, Patrick Esser, and Bj{\"o}rn Ommer.
\newblock High-resolution image synthesis with latent diffusion models.
\newblock In \emph{Proceedings of the IEEE/CVF Conference on Computer Vision and Pattern Recognition}, pages 10684--10695, 2022.

\bibitem[Ross and Doll{\'a}r(2017)]{focal}
Tsung-Yi Lin Priya~Goyal Ross and Girshick Kaiming He~Piotr Doll{\'a}r.
\newblock Focal loss for dense object detection.
\newblock In \emph{Proceedings of the IEEE International Conference on Computer Vision}, page 2980–2988, 2017.

\bibitem[Shen et~al.(2024)Shen, Hao, He, Zhao, Zhang, Liu, Bao, and Ding]{shen2024tempme}
Leqi Shen, Tianxiang Hao, Tao He, Sicheng Zhao, Yifeng Zhang, Pengzhang Liu, Yongjun Bao, and Guiguang Ding.
\newblock Tempme: Video temporal token merging for efficient text-video retrieval.
\newblock \emph{arXiv preprint arXiv:2409.01156}, 2024.

\bibitem[Shen et~al.(2025{\natexlab{a}})Shen, Gong, He, Zhang, Liu, Zhao, and Ding]{shen2025fastvid}
Leqi Shen, Guoqiang Gong, Tao He, Yifeng Zhang, Pengzhang Liu, Sicheng Zhao, and Guiguang Ding.
\newblock Fastvid: Dynamic density pruning for fast video large language models.
\newblock \emph{arXiv preprint arXiv:2503.11187}, 2025{\natexlab{a}}.

\bibitem[Shen et~al.(2025{\natexlab{b}})Shen, He, Gong, Yang, Zhang, Liu, Zhao, and Ding]{shen2025llava}
Leqi Shen, Tao He, Guoqiang Gong, Fan Yang, Yifeng Zhang, Pengzhang Liu, Sicheng Zhao, and Guiguang Ding.
\newblock Llava-mlb: Mitigating and leveraging attention bias for training-free video llms.
\newblock \emph{arXiv preprint arXiv:2503.11205}, 2025{\natexlab{b}}.

\bibitem[Sun et~al.(2024)Sun, Wang, Yu, Cui, Zhang, Zhang, and Wang]{sun2024eva}
Quan Sun, Jinsheng Wang, Qiying Yu, Yufeng Cui, Fan Zhang, Xiaosong Zhang, and Xinlong Wang.
\newblock Eva-clip-18b: Scaling clip to 18 billion parameters.
\newblock \emph{arXiv preprint arXiv:2402.04252}, 2024.

\bibitem[Tam et~al.(2024)Tam, Wu, Tsai, Lin, Lee, and Chen]{speakfree}
Zhi~Rui Tam, Cheng-Kuang Wu, Yi-Lin Tsai, Chieh-Yen Lin, Hung-yi Lee, and Yun-Nung Chen.
\newblock Let me speak freely? a study on the impact of format restrictions on performance of large language models.
\newblock \emph{arXiv preprint arXiv:2408.02442}, 2024.

\bibitem[Tan et~al.(2024)Tan, Yang, Qin, Yang, Zhang, and Li]{evalalign}
Zhiyu Tan, Xiaomeng Yang, Luozheng Qin, Mengping Yang, Cheng Zhang, and Hao Li.
\newblock Evalalign: Evaluating text-to-image models through precision alignment of multimodal large models with supervised fine-tuning to human annotations.
\newblock \emph{arXiv preprint arXiv:2406.16562}, 2024.

\bibitem[Vaswani(2017)]{vaswani2017attention}
A Vaswani.
\newblock Attention is all you need.
\newblock \emph{Advances in Neural Information Processing Systems}, 2017.

\bibitem[Wang et~al.(2024{\natexlab{a}})Wang, Chen, Lin, Han, and Ding]{wang2024repvit}
Ao Wang, Hui Chen, Zijia Lin, Jungong Han, and Guiguang Ding.
\newblock Repvit: Revisiting mobile cnn from vit perspective.
\newblock In \emph{Proceedings of the IEEE/CVF Conference on Computer Vision and Pattern Recognition}, pages 15909--15920, 2024{\natexlab{a}}.

\bibitem[Wang et~al.(2024{\natexlab{b}})Wang, Chen, Liu, Chen, Lin, Han, et~al.]{wang2024yolov10}
Ao Wang, Hui Chen, Lihao Liu, Kai Chen, Zijia Lin, Jungong Han, et~al.
\newblock Yolov10: Real-time end-to-end object detection.
\newblock \emph{Advances in Neural Information Processing Systems}, 37:\penalty0 107984--108011, 2024{\natexlab{b}}.

\bibitem[Wang et~al.(2021)Wang, Xiang, Yang, Qian, Hu, Huang, Han, Guo, and Ding]{wang2021lode}
Zerun Wang, Liuyu Xiang, Fan Yang, Jinzhao Qian, Jie Hu, Haidong Huang, Jungong Han, Yuchen Guo, and Guiguang Ding.
\newblock Lode: Deep local deblurring and a new benchmark.
\newblock \emph{arXiv preprint arXiv:2109.09149}, 2021.

\bibitem[Wei et~al.(2022)Wei, Wang, Schuurmans, Bosma, Xia, Chi, Le, Zhou, et~al.]{wei2022chain}
Jason Wei, Xuezhi Wang, Dale Schuurmans, Maarten Bosma, Fei Xia, Ed Chi, Quoc~V Le, Denny Zhou, et~al.
\newblock Chain-of-thought prompting elicits reasoning in large language models.
\newblock \emph{Advances in Neural Information Processing Systems}, 35:\penalty0 24824--24837, 2022.

\bibitem[Wu et~al.(2023{\natexlab{a}})Wu, Zhang, Zhang, Chen, Liao, Wang, Li, Sun, Yan, Zhai, et~al.]{q-bench}
Haoning Wu, Zicheng Zhang, Erli Zhang, Chaofeng Chen, Liang Liao, Annan Wang, Chunyi Li, Wenxiu Sun, Qiong Yan, Guangtao Zhai, et~al.
\newblock Q-bench: A benchmark for general-purpose foundation models on low-level vision.
\newblock \emph{arXiv preprint arXiv:2309.14181}, 2023{\natexlab{a}}.

\bibitem[Wu et~al.(2023{\natexlab{b}})Wu, Zhang, Zhang, Chen, Liao, Li, Gao, Wang, Zhang, Sun, et~al.]{qalign}
Haoning Wu, Zicheng Zhang, Weixia Zhang, Chaofeng Chen, Liang Liao, Chunyi Li, Yixuan Gao, Annan Wang, Erli Zhang, Wenxiu Sun, et~al.
\newblock Q-align: Teaching lmms for visual scoring via discrete text-defined levels.
\newblock \emph{arXiv preprint arXiv:2312.17090}, 2023{\natexlab{b}}.

\bibitem[Wu et~al.(2024{\natexlab{a}})Wu, Zhang, Zhang, Chen, Liao, Wang, Xu, Li, Hou, Zhai, et~al.]{q-instruct}
Haoning Wu, Zicheng Zhang, Erli Zhang, Chaofeng Chen, Liang Liao, Annan Wang, Kaixin Xu, Chunyi Li, Jingwen Hou, Guangtao Zhai, et~al.
\newblock Q-instruct: Improving low-level visual abilities for multi-modality foundation models.
\newblock In \emph{Proceedings of the IEEE/CVF Conference on Computer Vision and Pattern Recognition}, pages 25490--25500, 2024{\natexlab{a}}.

\bibitem[Wu et~al.(2025)Wu, Zhu, Zhang, Zhang, Chen, Liao, Li, Wang, Sun, Yan, et~al.]{wu2025towards}
Haoning Wu, Hanwei Zhu, Zicheng Zhang, Erli Zhang, Chaofeng Chen, Liang Liao, Chunyi Li, Annan Wang, Wenxiu Sun, Qiong Yan, et~al.
\newblock Towards open-ended visual quality comparison.
\newblock In \emph{European Conference on Computer Vision}, pages 360--377. Springer, 2025.

\bibitem[Wu et~al.(2024{\natexlab{b}})Wu, Ma, Liang, Yang, and Zhang]{wu2024comprehensive}
Tianhe Wu, Kede Ma, Jie Liang, Yujiu Yang, and Lei Zhang.
\newblock A comprehensive study of multimodal large language models for image quality assessment.
\newblock \emph{arXiv preprint arXiv:2403.10854}, 2024{\natexlab{b}}.

\bibitem[Wu et~al.(2023{\natexlab{c}})Wu, Sun, Zhu, Zhao, and Li]{human_preference_score}
Xiaoshi Wu, Keqiang Sun, Feng Zhu, Rui Zhao, and Hongsheng Li.
\newblock Human preference score: Better aligning text-to-image models with human preference.
\newblock In \emph{Proceedings of the IEEE/CVF International Conference on Computer Vision}, pages 2096--2105, 2023{\natexlab{c}}.

\bibitem[Xu et~al.(2024)Xu, Liu, Wu, Tong, Li, Ding, Tang, and Dong]{imagereward}
Jiazheng Xu, Xiao Liu, Yuchen Wu, Yuxuan Tong, Qinkai Li, Ming Ding, Jie Tang, and Yuxiao Dong.
\newblock Imagereward: Learning and evaluating human preferences for text-to-image generation.
\newblock \emph{Advances in Neural Information Processing Systems}, 36, 2024.

\bibitem[Yang et~al.(2022)Yang, Xu, Chen, Guo, Han, Ni, and Ding]{yang2022ground}
Fan Yang, Xinhao Xu, Hui Chen, Yuchen Guo, Jungong Han, Kai Ni, and Guiguang Ding.
\newblock Ground plane matters: Picking up ground plane prior in monocular 3d object detection.
\newblock \emph{arXiv preprint arXiv:2211.01556}, 2022.

\bibitem[Yang et~al.(2023)Yang, Xu, Chen, Guo, He, Ni, and Ding]{yang2023gpro3d}
Fan Yang, Xinhao Xu, Hui Chen, Yuchen Guo, Yuwei He, Kai Ni, and Guiguang Ding.
\newblock Gpro3d: Deriving 3d bbox from ground plane in monocular 3d object detection.
\newblock \emph{Neurocomputing}, 562:\penalty0 126894, 2023.

\bibitem[Yang et~al.(2024{\natexlab{a}})Yang, Chen, He, Zhao, Zhang, Ni, and Ding]{yang2024geometry}
Fan Yang, Hui Chen, Yuwei He, Sicheng Zhao, Chenghao Zhang, Kai Ni, and Guiguang Ding.
\newblock Geometry-guided domain generalization for monocular 3d object detection.
\newblock In \emph{Proceedings of the AAAI Conference on Artificial Intelligence}, pages 6467--6476, 2024{\natexlab{a}}.

\bibitem[Yang et~al.(2024{\natexlab{b}})Yang, Zhao, Zhang, Chen, Chen, Tang, Lu, Xu, Yang, Han, et~al.]{yang2024llmi3d}
Fan Yang, Sicheng Zhao, Yanhao Zhang, Haoxiang Chen, Hui Chen, Wenbo Tang, Haonan Lu, Pengfei Xu, Zhenyu Yang, Jungong Han, et~al.
\newblock Llmi3d: Empowering llm with 3d perception from a single 2d image.
\newblock \emph{arXiv preprint arXiv:2408.07422}, 2024{\natexlab{b}}.

\bibitem[You et~al.(2023)You, Li, Gu, Yin, Xue, and Dong]{you2023depicting}
Zhiyuan You, Zheyuan Li, Jinjin Gu, Zhenfei Yin, Tianfan Xue, and Chao Dong.
\newblock Depicting beyond scores: Advancing image quality assessment through multi-modal language models.
\newblock \emph{arXiv preprint arXiv:2312.08962}, 2023.

\bibitem[Zhang et~al.(2022)Zhang, Zhang, Li, and Smola]{zhang2022automatic}
Zhuosheng Zhang, Aston Zhang, Mu Li, and Alex Smola.
\newblock Automatic chain of thought prompting in large language models.
\newblock \emph{arXiv preprint arXiv:2210.03493}, 2022.

\bibitem[Zhang et~al.(2024)Zhang, Wu, Li, Zhou, Sun, Min, Chen, Liu, Lin, and Zhai]{A-Bench}
Zicheng Zhang, Haoning Wu, Chunyi Li, Yingjie Zhou, Wei Sun, Xiongkuo Min, Zijian Chen, Xiaohong Liu, Weisi Lin, and Guangtao Zhai.
\newblock A-bench: Are lmms masters at evaluating ai-generated images?
\newblock \emph{arXiv preprint arXiv:2406.03070}, 2024.

\end{thebibliography}
